\documentclass{article}
\usepackage{arxiv}
\usepackage{amsmath,amssymb}
\usepackage{algorithm, algpseudocode}
\usepackage{graphicx}
\usepackage{textcomp}
\usepackage{xcolor}
\usepackage{url}
\usepackage{hyperref}
\usepackage{booktabs}
\usepackage{arydshln}
\usepackage{cleveref}
\usepackage{multirow}
\usepackage{caption}
\usepackage{cite}

\makeatletter
\def\adl@drawiv#1#2#3{%
        \hskip.5\tabcolsep
        \xleaders#3{#2.5\@tempdimb #1{1}#2.5\@tempdimb}%
                #2\z@ plus1fil minus1fil\relax
        \hskip.5\tabcolsep}
\newcommand{\cdashlinelr}[1]{%
  \noalign{\vskip\aboverulesep
           \global\let\@dashdrawstore\adl@draw
           \global\let\adl@draw\adl@drawiv}
  \cdashline{#1}
  \noalign{\global\let\adl@draw\@dashdrawstore
           \vskip\belowrulesep}}
\makeatother

\newcommand{\R}{\mathbb{R}}
\newcommand{\x}{\boldsymbol{x}}
\newcommand{\y}{\boldsymbol{y}}
\newcommand{\X}{\boldsymbol{X}}
\newcommand{\Y}{\boldsymbol{Y}}

\makeatletter
\newcommand{\linebreakand}{%
  \end{@IEEEauthorhalign}
  \hfill\mbox{}\par
  \mbox{}\hfill\begin{@IEEEauthorhalign}
}
\makeatother

\title{CLeAN: \textbf{C}ontinual \textbf{Le}arning \textbf{A}daptive \textbf{N}ormalization in Dynamic Environments}

\author{
Isabella Marasco \\
  Department of Computer Science and Engineering\\
  University of Bologna\\
  Bologna, 40126, Italy \\
  \texttt{isabella.marasco4@unibo.it} 
   \And
 Davide Evangelista \\
  Department of Computer Science and Engineering\\
  University of Bologna\\
  Bologna, 40126, Italy \\
  \texttt{davide.evangelista5@unibo.it}
  \And
 Elena Loli Piccolomini \\
  Department of Computer Science and Engineering\\
  University of Bologna\\
  Bologna, 40126, Italy \\
  \texttt{elena.loli@unibo.it} 
 \And
 Michele Colajanni\\
  Department of Computer Science and Engineering\\
  University of Bologna\\
  Bologna, 40126, Italy \\
  \texttt{michele.colajanni@unibo.it} \\
}
\date{}
\begin{document}
\maketitle

%%%%%%%%%%%%%%%%%%%%%%%%%%%%%%%%%%%%%%%%%%%
% ABSTRACT
%%%%%%%%%%%%%%%%%%%%%%%%%%%%%%%%%%%%%%%%%%%
\begin{abstract}
% The rapid evolution of network environments present significant challenges for Intrusion Detection Systems (IDS). While Artificial Intelligence-based IDS have demonstrated notable success, their reliance on static data distributions imposes limitations on their adaptability to dynamic, real-world scenarios. 
Artificial intelligence systems predominantly rely on static data distributions, making them ineffective in dynamic real-world environments, such as cybersecurity, autonomous transportation, or finance, where data shifts frequently. Continual learning offers a potential solution by enabling models to learn from sequential data while retaining prior knowledge. However, a critical and underexplored issue in this domain is data normalization. Conventional normalization methods, such as min-max scaling, presuppose access to the entire dataset, which is incongruent with the sequential nature of continual learning. In this paper we introduce Continual Learning Adaptive Normalization (CLeAN), a novel adaptive normalization technique designed for continual learning in tabular data. CLeAN involves the estimation of global feature scales using learnable parameters that are updated via an Exponential Moving Average (EMA) module, enabling the model to adapt to evolving data distributions. Through comprehensive evaluations on two datasets and various continual learning strategies, including Resevoir Experience Replay, A-GEM, and EwC we demonstrate that CLeAN not only improves model performance on new data but also mitigates catastrophic forgetting. The findings underscore the importance of adaptive normalization in enhancing the stability and effectiveness of tabular data, offering a novel perspective on the use of normalization to preserve knowledge in dynamic learning environments.
\end{abstract}


%%%%%%%%%%%%%%%%%%%%%%%%%%%%%%%%%%%%%%%%%%%
% SECTIONS
%%%%%%%%%%%%%%%%%%%%%%%%%%%%%%%%%%%%%%%%%%%
\section{Introduction}
\label{introduction}
Artificial Intelligence (AI) has demonstrated remarkable success across numerous domains, achieving state-of-the-art performance on increasingly complex tasks. However, this success has been primarily realized under a critical assumption: that data distributions remain stable over time. This is rarely holds in real-world applications. In domains such as cybersecurity~\cite{yang2022systematic}, autonomous transportation~\cite{wakchaure2023application}, finance~\cite{ramjattan2024comparative}, or healthcare~\cite{ rong2020artificial}, data distributions evolve continuously due to emerging attack vectors, shifting market dynamics, changing environmental conditions, and evolving patient populations. These temporal shifts in data distribution pose a fundamental challenge to realize AI systems, often leading to performance degradation when models encounter patterns that deviate from their training data.
Consequently, the traditional \emph{"train once and use forever"} paradigm is often inadequate, as these environments are characterized by continuous shifts in statistical data distributions.
To cope with this issue, continual learning emerges as an approach that allows learning from dynamic data distribution without requiring re-training from scratch.
Continual learning~\cite{wang2024comprehensive, parisi2019continual}, also referred to as lifelong learning, is the capacity of a model to acquire knowledge from a continuous stream of data without forgetting previously acquired knowledge. In contrast to conventional machine learning, which presupposes the availability of all data at once, continual learning considers the challenge of learning when new data is presented sequentially, possibly with variation in their distribution. Continual learning includes several scenarios~\cite{van2022three}, such as Task Incremental Learning~\cite{oren2021defense}, in which task identity is known at inference, Domain Incremental Learning~\cite{mirza2022efficient}, in which data distribution shifts for existing classes, Class Incremental Learning~\cite{zhou2024class}, which required to acquire new classes over time and differentiate them from all previously encountered classes without explicit task identifiers, and Instance-Incremental Learning~\cite{nie2024decision}, where the data distribution shifts gradually within the same task and arrive in batches. In this paper, we focus on this latter scenario, which most closely reflects the continuous evolution of data distributions observed in the real-world.

In order to address the limitation of traditional machine learning, some studies have explored adaptation of classical methods to continual learning, such as the Incremental Random Forest~\cite{lakshminarayanan2014mondrian} Online Gradient Boosting~\cite{beygelzimer2015online}, and Online AdaBoost~\cite{santos2020online}. Despite the great performance shown by these algorithms, their use is still limited as they come with the drawback that the amount of memory required to store the parameters grows exponentially, making them unusable even for relatively small-scale problems. More recently, neural networks have emerged as a promising solution to this challenge, as the iterative nature of the Stochastic Gradient Descent (SGD)-like algorithms commonly used for the training of these models naturally enables incremental updates to parameters as new data streams are introduced.

However, a challenge in this context is \emph{catastrophic forgetting}~\cite{de2021continual, chen2018continual}, which refers to the phenomenon where the performance of a model degrades after certain patterns are no longer encountered. To address this challenge, various strategies have been proposed in the literature, including \emph{replay-based approaches}~\cite{bagus2021investigation, rolnick2019experience}, that aim to approximate the distribution of previous data either by storing a small subset of old training samples in a buffer, or by employing a generative model to generate synthetic samples, \emph{optimization-based approaches}~\cite{chaudhry2018efficient, lopez2017gradient} that directly manipulate the optimization process to mitigate catastrophic forgetting and promote knowledge transfer, and \emph{regularization-based approaches}~\cite{kann2023evaluation, nokhwal2023rtra}, where the idea is to add explicit regularization terms to the loss function, encouraging the model to retain important information from previous tasks while learning new ones.

An important step that becomes particularly prominent when employing AI-based approaches is \emph{data normalization}. In general, normalization of tabular data standardizes the range of feature values, preventing features with larger numerical values from disproportionately influencing the model and promoting more balanced and effective learning~\cite{umar2025effects, marasco2025continual, yaras2022neural}.

In dynamic environments, such as cybersecurity, where data distributions evolve due to the emergence of threats and system modifications, conventional normalization techniques become less effective. This distribution shift can result in inconsistencies in feature scaling between previously encountered and newly observed data, which can ultimately degrade model performance and destabilize learned representations. Therefore, the realization of adaptive normalization strategies which follows the evolution of data distributions is imperative. 
Despite its significance, this open issue remains unexplored in the literature. \\
In this paper, we initiate a systematic study to establish a structured framework for addressing the normalization challenge in tabular data within the continual learning context. First, we demonstrate that data normalization is a critical issue by experimentally showing how different normalization schemes significantly impact the performance of a trained AI-based system. Specifically, we compare the accuracy of a fixed neural network model trained on datasets normalized using several basic techniques selected as baselines, which consider only present and past data. These are contrasted with \emph{global normalization}, a technique that normalizes the entire dataset by considering present, past, and future data~\cite{amalapuram2024soul, channappayya2023augmented}. While global normalization consistently outperforms the other methods, it is not feasible in practice because it relies on future information that is not available in real-time scenarios. In this initial analysis, we therefore focus on two alternatives: \emph{local normalization}, where each data chunk is scaled using its own minimum and maximum values~\cite{talpini2024hierarchical}, and \emph{continual normalization}~\cite{pham2022continual}, an adaptive normalization strategy originally developed for computer vision tasks and adapted here to our continual-learning setup. 

We then introduce a novel learned normalization scheme, \emph{\textbf{C}ontinual \textbf{Le}arning \textbf{A}daptive \textbf{N}ormalization (CLeAN)}, which differs from local normalization in that it employs learnable parameters to estimate the global minimum and maximum values of the data based on the current chunk. To mitigate instabilities arising from rapid shifts in data distribution, these parameters are updated using an Exponential Moving Average (EMA) approach. Extensive experiments across multiple datasets and continual learning strategies such as Reservoir Experience Replay (RER)~\cite{vitter1985random,rolnick2019experience}, Average Gradient Episodic Memory (A-GEM)~\cite{chaudhry2018efficient}, and Elastic Weight Consolitation (EWC)~\cite{kirkpatrick2017overcoming} demonstrate that our normalization method not only significantly improves model performance on new data but also enhances memory retention of past data. Remarkably, this retention occurs even without explicit regularization. 
This discovery introduces a new paradigm for addressing catastrophic forgetting, leveraging data normalization as a mechanism for preserving information from previous experiences.

The remainder of this paper is organized as follows. In~\Cref{related_work}, we review related works, emphasizing how normalization is addressed in the current literature. In~\Cref{Methodology}, we introduce our proposed method, discussing the limitations of classical normalization techniques and explaining why our approach is expected to outperform them. In~\Cref{ExperimentalResults}, we detail our experimental setup, focusing on dataset selection, the buffering techniques employed, the competing normalization methods, and we present the results. Finally, in~\Cref{conclusions}, we summarize the key findings and highlight the main take-home messages from our experiments.

\section{Related Work}
\label{related_work}

Continual learning has gained significant traction in robotics~\cite{lesort2020continual, churamani2020continual, rebuffi2017icarl}, image classification~\cite{ermis2022continual, lomonaco2022cvpr, qu2021recent}, and video or image-based anomaly detection~\cite{rostami2021detection, doshi2020continual}.
However, its application to tabular data remains limited, especially regarding the critical issue of data normalization in dynamic environments. To the best of our knowledge, no prior work systematically addresses normalization in continual learning for tabular data. 
One line of work that touches on normalization in continual learning is Continual Normalization (CN)~\cite{pham2022continual}, which revisits Batch Normalization (BN) for online continual learning in vision tasks. CN addresses statistical bias in BN by modifying how normalization statistics are accumulated during training, improving retention and mitigating forgetting across tasks. Although the original formulation of CN is tailored to deep convolutional networks, it can be naturally adapted to tabular data, and we leverage this adaptation in our experimental setting.

Other works related to catastrophic forgetting in tabular data primarily considers domain-specific applications rather than general continual learning frameworks. 
For example, in cybersecurity Prasath Sai et al.~\cite{prasath2022analysis} study data distribution drift in deep learning–based intrusion detection systems, showing how covariate shift can induce catastrophic forgetting. They propose an eight-stage framework to quantify distributional changes and apply min–max normalization on the NSL-KDD and CICIDS-2017 datasets.
Channappayya et al.~\cite{channappayya2023augmented} extend the Class Balancing Reservoir Sampling (CBRS) to manage class imbalance and use Perturbation Assistance for Parameter Approximation (PAPA) to reduce computational overhead, applying min-max normalization as a preprocessing step. 
In the financial domain, Rahaman et al.~\cite{ramjattan2024comparative} realize a hybrid normalization framework to handle multi-source data complexities. They employ min-max normalization to reduce magnitude disparities. While Li et al.~\cite{li2024doctor} address adaptability in wearable health monitoring via DOCTOR, a multi-headed DNN framework that leverages replay-based continual learning and integrates min-max normalization for effective data preprocessing.
In these articles, either the normalization process is not specified, or it is applied globally to the entire dataset. This latter approach implies the necessity of prior knowledge of the entire dataset, a principle that stands in contrast to the idea of continual learning where data streams in sequentially.
To address this issue, we propose an innovative normalization approach based on a neural network. This approach introduces a dynamic normalization mechanism that adapts to changing data stream distributions and can be applied to a wide range of continual learning scenarios involving tabular data.
Unlike static and a priori normalization techniques employed on the entire dataset, our method continuously adjusts the normalization process to align with the evolving data distribution. 

Talpini J. et al. in~\cite{talpini2024hierarchical} propose a novel hierarchical model for IDS using continual learning, where a Bayesian Neural Network classifies benign and malicious traffic, and a generative multiclass classifier incrementally detects new attacks. In~\cite{marasco2025continual} the authors integrate continual learning into a system that predicts vessel trajectories. In these papers, they standardize the data per task by ensuring each feature has zero mean and unit variance, it is susceptible to instability due to rapid shifts in data distribution. In contrast, our proposed method combines dynamic normalization, neural network-based, with an EMA, which mitigates the effects of abrupt distribution shifts. The EMA smooths updates during periods of significant change, ensuring the model adapts without compromising previously acquired knowledge.
\section{Methodology}\label{Methodology}
% Continual Learning
% Catastrofic Forgetting + Buffer
% Normalization
% Proposal

In this section, we introduce continual learning and formally define catastrophic forgetting, the primary challenge associated with it. Next, we discuss the normalization of tabular data and its relevance to continual learning, proposing a series of baseline techniques based on existing literature. Finally, we present our approach, CLeAN, designed to enhance the performance of neural network models and mitigate catastrophic forgetting, particularly in the presence of unexpectedly rapid distribution shifts, such as those occurring after network protocol changes or during system attacks.

\subsection{Notations}
Throughout this article, we will make use of some notations, which we report here for completeness. In particular, we denote the input dataset as $\X \in \R^{N \times d}$, whose dimensions $N$ and $d$ represent the number of datapoints in $\X$ and the number of features, respectively. Similarly, $\Y \in \R^{N \times K}$ represents the output dataset associated with $\X$, where $K \in \mathbb{N}$ is the number of classes. Moreover, we denote as $\X_{t:(t+k)} \in \R^{k \times d}$ the sub-dataset of $\X$ obtained by extracting the rows with index from $t$ to $t+k$. Similarly, $\x_t \in \R^d$ represents the $t$-th row of $\X$, so that $\X_{t : (t+k)} = [\x_t, \dots, \x_{t+k}]^T$. To simplify the notation, all the operations on $\X$ are assumed to be applied row-wise. For example, $\X - \x_t$ is the matrix obtained by subtracting $\x_t$ from each row of $\X$, while $\max(\X)$ is a $d$-dimensional vector whose $i$-th element contains the maximum on $\X$ along the $i$-th feature. We also denote as $\boldsymbol{0}$ and $\boldsymbol{1}$ the $d$-dimensional vectors of all zeros and all ones, respectively. Finally, whenever $A$ is a finite set, we denote as $| A |$ the cardinality of $A$, i.e. the number of elements contained in $A$.  

\subsection{Continual Learning}
Continual learning, also referred to as lifelong learning, is a training paradigm for machine learning models specifically designed for dynamic systems where data is continuously collected from an evolving stream, potentially altering its structure over time. It is based on two key assumptions:
\begin{enumerate}
    \item The training algorithm has access to only a limited amount of data at each time step, which becomes inaccessible once new data arrives, except for a pre-determined set called the \emph{buffer}, which helps retain past information as new data flows in.
    \item The data distribution changes over time, either due to natural fluctuations, intentional external attacks, or protocol modifications. These distribution shifts can be categorized into \emph{domain shifts}, where changes occur in the input data distribution, and \emph{class shifts}, where the set of possible output labels changes due to the emergence of new attacks. 
\end{enumerate}

More formally, let $\{ \X_t \}_{t \in [0, \infty)}$ be a sequence of datasets of shape $N_t \times d$, where $N_t$ represents the number of data points in $\X_t$, and $d$ is the (fixed) number of features. In the continual learning literature, each $\X_t$ is often referred to as an \emph{experience}, and $\{ \X_t \}_{t \in [0, \infty)}$ is the \emph{set of experiences} for the given training task.

Let $ f_\Theta: \mathbb{R}^d \to [0,1]^K $ be a machine learning model that maps an input $ \x_i^t \in \X_t $ to a probability vector $ \y_i^t := f_\Theta(\x_i^t) $, where $K$ denotes the number of possible classes. In this paper, we primarily focus on binary classification and therefore assume $K=2$, although the proposed method can be straightforwardly extended to the multiclass setting ($K>2$). In the binary case, the model output can be represented by a single probability value in $[0,1]$, corresponding to the likelihood that the input belongs to the positive class.

Given a loss function $\ell: \mathbb{R} \times \mathbb{R} \to \mathbb{R}_{\geq 0}$, the standard continual learning paradigm assumes that, at each time step $t$, the model parameters $\Theta$ are trained to minimize the expected loss:

\begin{equation} \label{eq:continual_learning_classic}
    \Theta_t \in \arg\min_{\Theta} \mathbb{E}_{(\x_i^t, \y_i^t) \sim (\X_t, \Y_t)} \left[ \ell (f_\Theta(\x_i^t), \y_i^t) \right],
\end{equation}
where $\Y_t \in \mathbb{R}^{N_t}$ is the vector of class labels corresponding to $\X_t$. This formulation implies that at each time step $t$, the model parameters are optimized solely on the available data, with no access to past ($t' < t$) or future ($t' > t$) information. This restriction becomes problematic if the distribution of $\X_t$ significantly differs from that of previous experiences $\X_{t'}$, as the model’s performance may degrade when previously seen data reappears. Later in this section, we will discuss strategies to address this limitation in practice. \\

A crucial aspect of continual learning is the choice of the machine learning model $ f_\Theta $. In traditional applications where data is assumed to be static, meaning the whole set of data is available at once, Random Forest (RF) is tipically used. However, RF is not well-suited for continual learning, because its training is \emph{direct}: given a dataset, the model computes optimal parameters in a single pass. When applied to continual learning, this results in parameters that are independent across time steps, causing the model to \emph{forget} previously learned information as soon as new data arrives, exacerbating the limitations described earlier.

Neural networks, in contrast, are typically trained using Stochastic Gradient Descent (SGD), where parameters are iteratively updated as:

\begin{equation}
    \Theta^{(k+1)} = \Theta^{(k)} - \nu_k \nabla_{\Theta} \ell(f_{\Theta^{(k)}} (\x_i^t), \y_i^t).
\end{equation}

This iterative nature makes neural networks naturally suited for continual learning, as training can simply \emph{continue} from where it left off when the dataset updates from $\X_t$ to $\X_{t+1}$. This allows the model to retain \emph{memory} of previously encountered samples, helping maintain performance as the data distribution evolves. However, over long time intervals, this approach alone is insufficient. Model parameters gradually drift away from optimal values for earlier time steps, leading to a decline in performance, a phenomenon known as \emph{catastrophic forgetting}.

\subsection{Catastrophic forgetting}\label{CatastroficForgetting}
Catastrophic forgetting is the phenomenon in which a model trained on a dataset $\X_t$ at a given timestep $t > 0$ is unable to recognize data seen in the past (typically for $t' \ll t$), leading to performance degradation even when it performs well on the current data.

Formally, let $f_{\Theta_t}$ be the model trained on dataset $\X_t$, and let $a_t^{t'}$ denote the accuracy of $f_{\Theta_t}$ when tested on data sampled from $\X_{t'}$. The model is said to suffer from catastrophic forgetting if $a_t^{t'} \ll a_t^t$, for $t' \leq t$.

Since catastrophic forgetting is unavoidable in most continual learning models, various techniques have been developed in recent years to mitigate this issue, as discussed in \Cref{introduction}. Methods for addressing catastrophic forgetting reformulate \Cref{eq:continual_learning_classic} as:
\begin{align}\label{eq:continual_learning_CF}
    \Theta_t^* \in \arg\min_{\Theta} \>\> \max_{t' \leq t} \mathbb{E}_{(\X_{t'}, \Y_{t'})} [ \ell (f_\Theta(\x_i^{t'}), \y_i^{t'}) ],
\end{align}
which encapsulates the idea that the model should perform well not only on current data but also on previously encountered datasets. Clearly, solving \eqref{eq:continual_learning_CF} is not feasible in practice, causing any model to inevitably lose information about older data distributions over time. The process of retaining knowledge from past data distributions is referred to as \emph{memorization}.

Methods aimed at improving memorization typically seek to approximate \eqref{eq:continual_learning_CF} in a tractable manner. Among these approaches, a particularly successful line of work is represented by buffer-based methods. A buffer is a limited memory storage of size $B \in \mathbb{N}$ used to retain previously seen examples, which can be replayed during training to mitigate catastrophic forgetting. The primary difference between various buffer-based methods lies in how data is selected and stored over time. The underlying objective is to ensure that, when training on $\X_t$ using the buffer $\mathcal{B}$, the resulting optimization remains as close as possible to the solution of \Cref{eq:continual_learning_CF}. To achieve this, several techniques have been proposed over the years.

One of the simplest yet most widely adopted approaches is \emph{Reservoir Experience Replay (RER)}~\cite{vitter1985random}. At each timestep $t$, a subset of samples is uniformly drawn from $\X_t$ and stored in a fixed-capacity buffer, with older entries being replaced once the buffer is full. During training, mini-batches are formed by combining current data from $\X_t$ with samples replayed from the buffer. This replay mechanism encourages the model to preserve knowledge of previously observed data distributions while adapting to newly arriving samples.

An alternative to RER, also considered in this work, is \emph{A-GEM}~\cite{chaudhry2018efficient}. Similarly to experience replay, A-GEM maintains a buffer of past examples; however, the buffer is exploited differently during training. Rather than replaying stored samples, A-GEM uses the buffer to constrain parameter updates so as to prevent interference with previously learned knowledge. Specifically, the gradient $\boldsymbol{g}_t^{(k)}$ computed on samples from $\X_t$ is projected onto the gradient $\boldsymbol{g}_{\mathcal{B}}$ computed on the buffer whenever $\boldsymbol{g}_{\mathcal{B}}^\top \boldsymbol{g}_t^{(k)} < 0$, that is,
\begin{align}
	\begin{cases}
		\boldsymbol{g}_t^{(k)} \leftarrow \boldsymbol{g}_t^{(k)} - 
		\dfrac{\boldsymbol{g}_{\mathcal{B}}^\top \boldsymbol{g}_t^{(k)}}{\boldsymbol{g}_{\mathcal{B}}^\top \boldsymbol{g}_{\mathcal{B}}}
		\boldsymbol{g}_{\mathcal{B}}, 
		& \text{if } \boldsymbol{g}_{\mathcal{B}}^\top \boldsymbol{g}_t^{(k)} < 0, \\
		\boldsymbol{g}_t^{(k)} \leftarrow \boldsymbol{g}_t^{(k)}, 
		& \text{otherwise}.
	\end{cases}
\end{align}
This projection enforces that updates on the current task do not increase the loss on previously observed data, thereby mitigating catastrophic forgetting.

Finally, we also consider \emph{Elastic Weight Consolidation (EWC)}~\cite{kirkpatrick2017overcoming}, a regularization-based approach that does not rely on an explicit memory buffer. EWC mitigates catastrophic forgetting by augmenting the training loss $\ell(f_\Theta(\x_i^t), \y_i^t)$ with a penalty term that discourages significant deviations from parameters learned in previous tasks. Concretely, this is achieved by introducing a quadratic regularizer that constrains the model parameters $\Theta$ to remain close to previously learned values $\Theta^{(t)}$, for instance through a term of the form $ \lVert \Theta - \Theta^{(t)} \rVert_2^2$, weighted according to their estimated importance.

While these methods have been shown to be effective in continual learning problems in computer vision \cite{li2024calibration,chaudhry2018efficient}, their performance on tabular data applications significantly depends on how data is normalized, as we will experimentally show in \Cref{ExperimentalResults}.

\subsection{Normalization} \label{Normalization}
Data normalization is a crucial step when training a Machine Learning model on tabular data, particularly when features exhibit significantly different scales. To understand why, consider the gradient of the loss function $\ell$ with respect to $ \Theta $:  
\begin{align}
\nabla_\Theta \ell(f_\Theta(\x_i^t), \y_i^t) = \frac{\partial \ell}{\partial f_\Theta} \cdot \nabla_\Theta f_\Theta(\x_i^t).
\end{align}
For any parameter $ \Theta_l $ associated with the $l$-th feature $\x_{i, l}^t$, the gradient term $\left(\nabla_\Theta f_\Theta(\x_i^t)\right)_l$ is proportional to $\x_{i, l}^t$. This implies that features with larger magnitudes lead to amplified gradients, resulting in larger parameter updates. Conversely, smaller-scale features yield smaller gradients and slower updates. Such imbalance can lead to unstable training, where the model disproportionately adapts to large-scale features while neglecting smaller ones. Normalizing the input ensures that all features contribute equally to the gradient, fostering stable and balanced learning. \\
Formally, we define a normalization algorithm as a function $ \mathcal{S}: \mathbb{R}^d \to \mathbb{R}^d $ such that the $l$-th feature of the normalized data $\hat{\x}_i^t$ depends only on the corresponding feature of the unnormalized input $ \x_i^t $, i.e., $ \left(\mathcal{S}(\x_i^t)\right)_l = \mathcal{S}(\x_{i, l}^t) $. Ideally, a normalization method should ensure that all features of $ \hat{\x}_i^t $ have comparable magnitudes, typically within $[0,1]$ or $[-1,1]$. 

Among various normalization techniques, we focus on min-max normalization, which is typically defined as:
\begin{align}\label{eq:minmax_classic}
    \hat{\x} = \mathcal{S}(\x) := \frac{\x - \boldsymbol{m}}{\boldsymbol{M} - \boldsymbol{m}},
\end{align}
where $ \boldsymbol{M} $ and $ \boldsymbol{m} $ denote the element-wise maximum and minimum of the dataset, respectively. While this method is standard in machine learning, caution is required in continual learning settings as, when data arrives sequentially, determining $ \boldsymbol{M} $ and $ \boldsymbol{m} $ becomes challenging. 

Many studies~\cite{amalapuram2024soul,channappayya2023augmented} assume access to the entire dataset and define:
\begin{align}
    \boldsymbol{M} := \max \left( \bigcup_{t \geq 0} \X_t \right), \quad
    \boldsymbol{m} := \min \left( \bigcup_{t \geq 0} \X_t \right).
\end{align}
While effective, this approach is impractical in real-time data collection since it requires access to the entire sequence $ \{ \X_t \}_{t \geq 0} $. Consequently, results obtained using this method should be interpreted with caution, as they overestimate the normalizer's ability and fail to account for adversarial scenarios where an attacker manipulates input scales to bypass security measures. Nevertheless, due to its widespread adoption, we include this method in our comparisons and we threat it as a theoretical upper bound in performance, referring to it as \emph{global normalization} since it leverages global data statistics. \\

A more practical alternative, which we name \emph{local normalization}, has appeared in only a few studies~\cite{talpini2024hierarchical}. Here, normalization is performed independently for each dataset $ \X_t $, using:
\begin{align}
    \begin{split}
        &\boldsymbol{M}_t := \max (\X_t), \quad \boldsymbol{m}_t := \min( \X_t), \\
        &\hat{\X}_t = \mathcal{S}_t (\X_t) := \frac{\X_t - \boldsymbol{m}_t}{\boldsymbol{M}_t - \boldsymbol{m}_t}.
    \end{split}
\end{align}
A key limitation of local normalization is that the same data point $ \x_i $ can be mapped to different values when observed in different datasets $ \X_t $ and $ \X_{t'} $, i.e.,
\begin{align}
    \hat{\x}_i^t = \mathcal{S}_t(\x_i) \neq \mathcal{S}_{t'} (\x_i) = \hat{\x}_{i}^{t'}.
\end{align}

This discrepancy implies that a model trained on $ \X_t $ to predict $ \hat{\x}_i^t $ may struggle to generalize when encountering $ \hat{\x}_{i}^{t'} $. This issue highlights how normalization impacts catastrophic forgetting: as we will demonstrate in \Cref{ExperimentalResults}, models trained with local normalization suffer from greater forgetting than those trained under the ideal global normalization framework.

Another drawback of local normalization is its sensitivity to sudden changes in the data distribution. If $ \boldsymbol{M}_{t+1} \gg \boldsymbol{M}_t $, previously normalized data points may undergo drastic rescaling, impairing model performance. This vulnerability can be exploited e.g., in cybersecurity settings, where an adversary may deliberately inject outliers into the data stream to induce abrupt rescaling effects, temporarily degrading detection performance and facilitating unauthorized access.

To address these limitations, adaptive schemes that update normalization statistics over time have been proposed, of which the most known is CN~\cite{amalapuram2024soul}. In CN, at each experience $\X_t$, the mean $\boldsymbol{\mu}_t$ and standard deviation $\boldsymbol{\sigma}_t$ of each feature are updated as
\begin{align}
\begin{split}
    &\boldsymbol{\mu}_t = (1-\lambda)\boldsymbol{\mu}_{t-1} + \lambda \, \mu(\X_t), \\
    &\boldsymbol{\sigma}_t = (1-\lambda)\boldsymbol{\sigma}_{t-1} + \lambda \, \sigma(\X_t),
\end{split}
\end{align}
where $\mu(\cdot)$ and $\sigma(\cdot)$ denote the per-feature mean and standard deviation computed on the current experience, and $\lambda$ controls the temporal smoothness of the update. The normalized data are then obtained through a standard $z$-score transformation:
\begin{align}
    \hat{\x}_i^t = \frac{\x_i^t - \boldsymbol{\mu}_t}{\boldsymbol{\sigma}_t + \epsilon}.
\end{align}
This formulation enables the model to gradually adapt to distributional changes without requiring access to previous experiences, offering a practical alternative to global normalization in streaming scenarios.  

However, CN relies exclusively on first- and second-order statistics computed independently for each feature and does not explicitly account for extreme values or rapid shifts in data ranges. As a result, it may still suffer from instability when the data distribution undergoes abrupt changes or exhibits heavy-tailed behavior: conditions that frequently arise in real-world tabular streams such as network traffic. These limitations motivate the need for a more robust and flexible normalization mechanism, capable of jointly adapting with the learning model while remaining resilient to sudden distribution shifts.

\begin{figure*}[tbh!]
    \centering
    \includegraphics[width=0.8\linewidth]{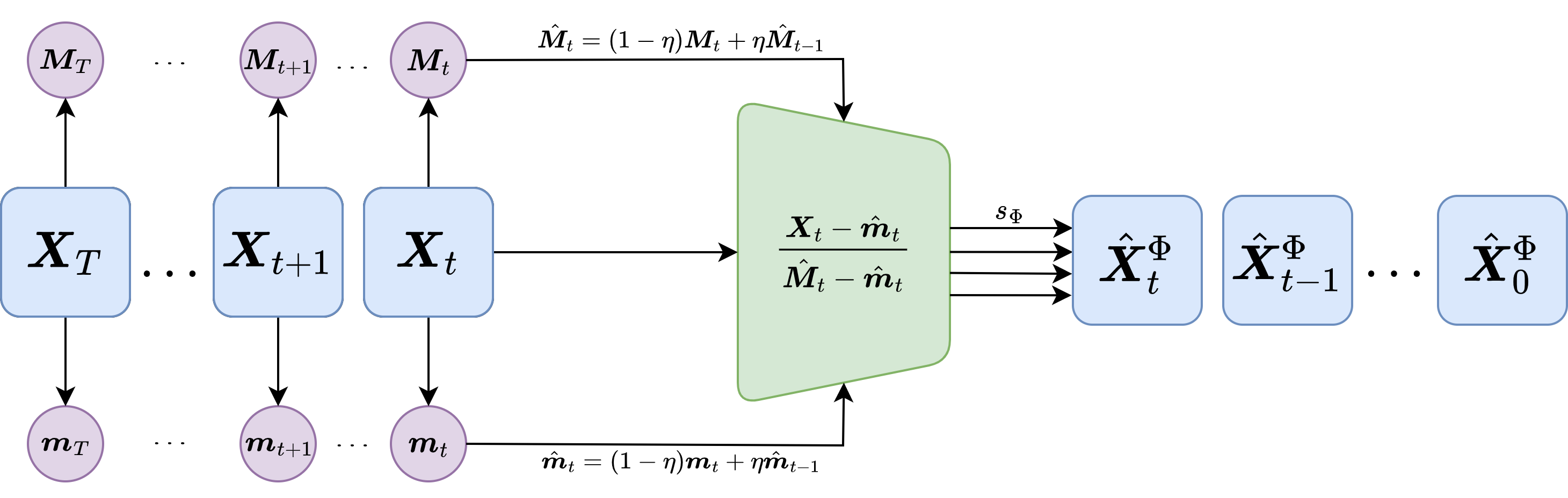}
    \caption{A graphical representation of our proposed normalization scheme: CLeAN.}
    \label{fig:CLeAN}
\end{figure*}

\subsection{Our proposal: CLeAN}
Motivated by the limitations of CN in handling abrupt range shifts and extreme values, we introduce CLeAN, a neural network–based normalization scheme designed for continual learning on tabular data. CLeAN estimates reliable normalization bounds $\boldsymbol{M}_t$ and $\boldsymbol{m}_t$ directly from the incoming data stream and adapts them over time by jointly training with the predictive model $f_\Theta$.
While CN adapts normalization using fixed-form, per-feature statistics, CLeAN introduces a learnable and range-aware normalization mechanism that evolves jointly with the model and remains robust to sudden distributional shifts.

More precisely, as depicted in \Cref{fig:CLeAN}, our proposed method is composed of two parts: a maximum/minimum estimation component, described by the $d$-dimensional vectors $\hat{\boldsymbol{M}}_t$ and $\hat{\boldsymbol{m}}_t$, and a learned scaling layer $s_\Phi: \R^d \to \R^d$, parameterized by $\Phi$, which is implemented by a neural network with diagonal weight matrix so that it satisfies the definition of a normalizer which should be applied element-wise. When new data $\X_t$ is given, its local maximum $\boldsymbol{M}_t = \max (\X_t)$ and minimum $\boldsymbol{m}_t = \min (\X_t)$ are computed and used to update the model estimators $\hat{\boldsymbol{M}}_t$ and $\hat{\boldsymbol{m}}_t$ via an Exponentially Moving Average scheme, as:
\begin{align}
\begin{split}
    \hat{\boldsymbol{M}}_t &= (1-\eta) \boldsymbol{M}_t + \eta \hat{\boldsymbol{M}}_{t-1}, \\
    \hat{\boldsymbol{m}}_t &= (1-\eta) \boldsymbol{m}_t + \eta \hat{\boldsymbol{m}}_{t-1},
\end{split}
\end{align}
where $\eta \approx 1$ is introduced to ensure the estimators are more resilient to quick transitions in the data statistics. Afterward, $\X_t$ gets min-max normalized via the estimated maximum and minimum as:
\begin{align}
    \hat{\X}_t = \frac{\X_t - \hat{\boldsymbol{m}}_t}{\hat{\boldsymbol{M}}_t - \hat{\boldsymbol{m}}_t},
\end{align}
and then adjusted by the scaling neural network $s_\Phi$ as:
\begin{align}
    \hat{\X}_t^\Phi = s_\Phi(\hat{\X}_t).
\end{align}
Finally, the normalized dataset $\hat{\X}_t^\Phi$ is plugged into the model $f_\Theta$ which is continuously trained on it (in both $\Phi$ and $\Theta$). The pseudo-code of this procedure is depicted in \Cref{alg:CLeAN}. \\

While the idea underlying CLeAN is conceptually simple, our experimental results show that the proposed method not only outperforms the same model trained with alternative normalization schemes, but also exhibits significantly improved memorization capabilities. These findings highlight the potential of adaptive normalization strategies to support continual learning in tabular settings, complementing existing techniques for mitigating catastrophic forgetting while remaining applicable in realistic streaming scenarios. In the next section, through extensive experiments on two benchmark datasets from the cybersecurity domain, we demonstrate the effectiveness and robustness of our approach.

\begin{algorithm}
\caption{The proposed CLeAN scheme}
\label{alg:CLeAN}
\begin{algorithmic}[1]
\Require A sequence $\{ \X_t \}_{t \in [0, \infty)}$ of experiences, an element-wise model $s_\Phi$, the EMA parameter $\eta > 0$

\Statex
\State Initialize $\hat{\boldsymbol{M}}_0 = \boldsymbol{1}$ and $\hat{\boldsymbol{m}}_0 = \boldsymbol{0}$
\For{$t > 0$}
    \State Compute $\boldsymbol{M}_t \leftarrow \max\left( \X_t \right)$ and $\boldsymbol{m}_t \leftarrow \min\left( \X_t \right)$
    \State Update $\hat{\boldsymbol{M}}_t$ and $\hat{\boldsymbol{m}}_t$ as: \begin{align*}
    \begin{split}
        \hat{\boldsymbol{M}}_t &\leftarrow (1-\eta) \boldsymbol{M}_t + \eta \hat{\boldsymbol{M}}_{t-1} \\
        \hat{\boldsymbol{m}}_t &\leftarrow (1-\eta) \boldsymbol{m}_t + \eta \hat{\boldsymbol{m}}_{t-1}
    \end{split}
    \end{align*}
    \State Compute: 
    \begin{align*}
        \hat{\X}_t \leftarrow \frac{\X_t - \hat{\boldsymbol{m}}_t}{\hat{\boldsymbol{M}}_t - \hat{\boldsymbol{m}}_t}
    \end{align*}
    \State Update $\hat{\X}_t^\Phi \leftarrow s_\Phi(\hat{\X}_t)$
    \State Train on $\hat{\X}_t^\Phi$ optimizing $\ell (f_\Theta(\hat{\X}_t^\Phi), \Y_t)$
\EndFor
\State \Return $(\hat{\boldsymbol{M}}_t, \hat{\boldsymbol{m}}_t, s_\Phi)$ \Comment{the trained CLeAN parameters}
\end{algorithmic}
\end{algorithm}
\section{Experimental results}\label{ExperimentalResults}
In this section, we describe the pre-processing, training, and evaluation strategies employed. Then, we present the experimental setup, including the rationale for each experiment. Lastly, we report and analyze the numerical results. All experiments were implemented using pure PyTorch without reliance on any other external deep learning frameworks. The code was executed on a MacBook equipped with an Apple M3 chip, utilizing its integrated GPU and CPU resources. While all computational choices are indicated below, a more detailed procedure is available in the project repository\footnote{Available after publication or under request}.

\subsection{Datasets}
To evaluate the effectiveness of the proposed CLeAN normalization technique, we focused our experimental validation on the cybersecurity domain, specifically Intrusion Detection Systems (IDS)~\cite{yang2022systematic}. An IDS is a system for monitoring network traffic with the aim of detecting malicious activities and security policy violations. However, the rapid growth of network traffic, coupled with evolving network architectures, has significantly increased the complexity required for these systems, underscoring the need for advanced solutions capable of adapting to sophisticated threats. Consequently, this domain represents an optimal scenario for validation of Continual Learning systems: network environments are inherently non-stationary, exhibiting abrupt feature distribution shifts driven by protocol updates, fluctuating traffic patterns, and the emergence of novel attack classes. 
For our benchmarks, we utilized the UNSW-NB15~\cite{moustafa2015unsw, moustafa2016evaluation, moustafa2017novel, moustafa2017big, sarhan2021netflow} and CICIDS-2017~\cite{sharafaldin2018toward} datasets, selected due to their frequent use in the literature on IDS and continual learning. Moreover, they are characterized by numerous features with a large range of values. This property is crucial to evaluate the efficacy of our proposed CLeAN scheme, which is designed to normalize data and manage significant distribution shifts dynamically, a requirement that is unique for numerical features.

The UNSW-NB15 dataset, realized by the University of New South Wales (UNSW), counts $2,540,047$ samples and $47$ features, including both numerical and categorical variables. Among these, nine features are categorical: \texttt{srcip}, \texttt{sport}, \texttt{dstip}, \texttt{dsport}, \texttt{proto}, \texttt{state}, \texttt{service}, \texttt{ct\_ftp\_cmd}, and \texttt{attack\_cat}. Therefore, to be able to process them into the neural network model, we need to convert them into numerical values. The \texttt{srcip} and \texttt{dstip}, representing the source and destination IP addresses, are each decomposed into four separate columns corresponding to their dot-separated octets (e.g. a value like \texttt{149.171.126.9} gets decomposed into four numerical features $[149, 171, 126, 9]$). The remaining categorical features are converted to numeric representations employing label encoding~\cite{poslavskaya2023encoding}. Following these preprocessing steps, the final dataset consists of $N = 2,540,047$ samples and $d = 53$ features, excluding the binary label. The dataset contains $2,218,764$ benign samples and a total of $321,283$ malicious samples distributed across these categories. 

The CICIDS-2017 dataset, provided by the Canadian Institute for Cybersecurity (CIC), consists of $2{,}830{,}743$ samples and $78$ numerical features, with no categorical variables. The following features exhibit constant values across all samples and are therefore removed: \texttt{Bwd PSH Flags}, \texttt{Fwd URG Flags}, \texttt{Bwd URG Flags}, \texttt{CWE Flag Count}, \texttt{Fwd Avg Bytes/Bulk}, \texttt{Fwd Avg Packets/Bulk}, \texttt{Fwd Avg Bulk Rate}, \texttt{Bwd Avg Bytes/Bulk}, \texttt{Bwd Avg Packets/Bulk}, and \texttt{Bwd Avg Bulk Rate}. Additionally, all records containing null values are discarded. After these preprocessing steps, the resulting dataset contains $N = 2{,}827{,}876$ samples and $d = 68$ features, excluding the label.

Both the UNSW-NB15 and CICIDS-2017 datasets exhibit substantial class imbalance, with benign samples forming the vast majority of instances. This imbalance increases the difficulty of training, as models must learn to recognize relatively rare attack patterns amid dominant normal traffic. Following common practice in IDS research, we frame the task as a binary classification problem: all attack types are grouped under a single \textit{attack} label ($1$), contrasted with the \textit{benign} label ($0$).

To simulate a data stream environment, each dataset is divided into sequential chunks. Specifically, we define chunks of fixed size $N_t = 500{,}000$ and synthetically generate experiences from contiguous portions of the full dataset $\X$ as
\begin{align}
\X_t := \X_{t N_t : (t+1)N_t}, \quad t = 0, 1, \dots
\end{align}
The set of experiences contains $T = 6$ elements for the UNSW-NB15 dataset and $T = 5$ for CICIDS-2017.
This design contrasts with the standard continual learning setup (e.g.,~\cite{zhang2024continual}), where data are first randomly partitioned into training and test sets before sequential tasks are derived from them. Our approach, where the natural data order is preserved during chunk generation, yields a more realistic emulation of real-world streaming scenarios.

\subsection{Experimental setup}
In our experiments, we investigate the effectiveness of CLeAN, comparing it to other normalization techniques discussed in~\Cref{Normalization}, such as \emph{global normalization}, \emph{local normalization}, and \emph{continual normalization}. We also consider three  methods to tame catastrophic forgetting: one where no buffer is used, which serves as a baseline, named \emph{fine-tuning} accordingly to the literature; an experience replay approach called \textit{reservoir experience replay}; an optimization-based method named \emph{A-GEM}; a regularization-based algorithm named \emph{EWC}. More details on the working mechanism of each of these methods can be found in~\Cref{CatastroficForgetting}.

To evaluate our methodology, we use a fully connected neural network with $4$ hidden layers, each with $128$ neurons, followed by a Rectified Linear Unit (ReLU) activation function. To avoid overfitting, dropout regularization with a probability of $p = 0.5$ is applied after each hidden layer. These parameters have been manually tuned to maximize the performance on the global normalization setup, which represents the theoretical upper bound for our methods. The output layer consists of a linear transformation followed by a sigmoid activation function, as it is usual for binary classification tasks. This process yields a value between $0$ and $1$, indicating the probability that the model assigns at the input to lie in the class $1$ (attack). This value is then converted to the actual prediction by a thresholding function $\tau: [0, 1] \to \{ 0, 1 \}$, defined as:
\begin{align}
    \tau\left( f_\Theta(\x_i^t) \right) = \begin{cases}
        $1$ &\mbox{ if } f_\Theta(\x_i^t) > \kappa, \\
        $0$ &\mbox{ otherwise.}
    \end{cases}
\end{align}
The thresholding value has been consistently set to $\kappa = 0.5$ throughout the experiments.
The network is then trained by minimizing the Binary Cross-Entropy loss function with the \texttt{Adam} optimizer for a total of $20$ epochs per experience and a batch size of $20,000$. The learning rate has been set to $10^{-3}$ for the USNW-NB15 dataset, and to $5 \times 10^{-4}$ for the CIC-IDS2017 dataset. These values have been \emph{heuristically} selected to ensure maximum performance.
The splitting of data streams for training and testing in continual learning, especially with tabular data, necessitates meticulous consideration to ensure meaningful evaluation. Some papers in literature present results derived from methodologies where the train and test partitioning or normalization procedures may not strictly match how a system would operate sequentially in reality. This can lead to performance figures that are not robust outside the experimental setup. In this paper we adopt a pragmatic evaluation scheme within each data chunk, focusing on practical applicability and realistic simulation. In particular, we divide each dataset into sequential chunks that represent the incoming data stream. Each chunk is further divided into a training set and a test set. During the training phase for each specific chunk, the data is normalized as they are used for training. The test set for each chunk remains in its original scale and is normalized only at the time of evaluation. The normalization scale applied to the test set is derived from the scale range that is actually applied to the training set at the current experience.
This evaluation strategy enables the assessment of the ability of the model to generalize to unseen data from the current data distribution, while reflecting the inherent challenges of learning from a non-stationary data stream.

\subsection{Metrics}\label{ssec:metrics}
%to modify for new metrics to use
To evaluate the effectiveness of the different normalization techniques and forgetting mitigation strategies, we use three common metrics: average accuracy, average forgetting, and average Area Under the Receiver Operating Characteristic Curve (AUROC). 
These metrics are chosen to quantify both the ability of the model to adapt to new data distributions (experiences) and its capacity to retain knowledge from previous experiences, thus measuring the extent of catastrophic forgetting.
%%%
Let $\{\X_t\}_{t=1}^{T}$ be the sequence of experiences, and let $f_{\Theta_t}$ denote the model after training sequentially on experiences $\X_1, \dots, \X_t$. We define $a_t^{t'}$ as the accuracy of model $f_{\Theta_t}$ on the test set of experience $\X_{t'}$, i.e.:  
\begin{align}\label{eq:accuracy}
    a_t^{t'} = \frac{\left| \{ \y_i^{t'} \in \Y_{t'} \mid \tau\left(f_{\Theta_t}(\x_i^{t'})\right) = \y_i^{t'} \} \right|}{N}
\end{align}
The \textit{accuracy $a_t^{t}$} measures performance on the test set of the current experience $\X_t$ after training on it, by reflecting adaptation to recent data.

Related to \Cref{eq:accuracy}, the \textit{average accuracy ($A_t$)}, provides an overall measure across all experiences up to time $t$ using model $f_{\Theta_t}$, gauging generalization and the balance between learning and retention. It is calculated as the average accuracy across all tasks as follows: 
\begin{align}
    A_t = \frac{1}{t} \sum_{t'=1}^{t} a_t^{t'}.
\end{align}

The \textit{Average Forgetting ($\Gamma_t$)}, introduced in \cite{chaudhry2019tiny}, quantifies catastrophic forgetting by measuring the average performance degradation on previous tasks after learning task $t$. It is defined as the average over all the experiences up to time $t$ of a quantity $\gamma_{t'}^{t}$ named \emph{forgetting}. It quantifies the maximum drop in accuracy observed on the test set of each $\X_{t'}$ for $t' < t$ compared to the accuracy on $\X_t$ when the model is trained on $\X_t$, that is, 
\begin{align}
    \gamma_{t}^{t'} = \max_{s \in \{1, ..., t-1\}} a_s^{t'} - a_t^{t'}.
\end{align}
Consequently, the \textit{Average forgetting ($\Gamma_t$}) is computed as:
\begin{align}
    \Gamma_t = \frac{1}{t-1} \sum_{t'=1}^{t-1} \gamma_{t}^{t'}.
\end{align}
A lower average forgetting value indicates better mitigation of catastrophic forgetting. Together, these two metrics provide a comprehensive evaluation of how well a continual learning model navigates the stability-plasticity dilemma inherent in learning from evolving data streams.

Finally, we evaluate the average AUROC, defined at experience $t$ as the mean of the AUROC values computed over all experiences up to $t$. In general,  AUROC measures the model’s ability to distinguish between benign and malicious samples across varying decision thresholds. It is especially informative in imbalanced settings, as it reflects the model’s ranking performance independently of the chosen threshold. A higher AUROC indicates that the model tends to assign greater confidence to positive (attack) instances than to negative (benign) ones even when the final classification threshold is suboptimal. By reporting AUROC, we complement standard accuracy-based metrics with a robust measure of detection reliability and stability under class imbalance. For a comprehensive discussion of the AUROC computation and its theoretical properties, see \cite{fawcett2006introduction}.

\begin{table*}[tbh]
    \centering
    \renewcommand{\arraystretch}{1.2}
    \setlength{\tabcolsep}{4pt}
    \begin{tabular}{ccccc}
        & \texttt{Finetuning} & \texttt{Reservoir} & \texttt{A-GEM} & \texttt{EwC} \\
        \rotatebox{90}{\hspace{18px}\textbf{Average Accuracy ($\uparrow$)}} &
        \includegraphics[width=0.22\linewidth]{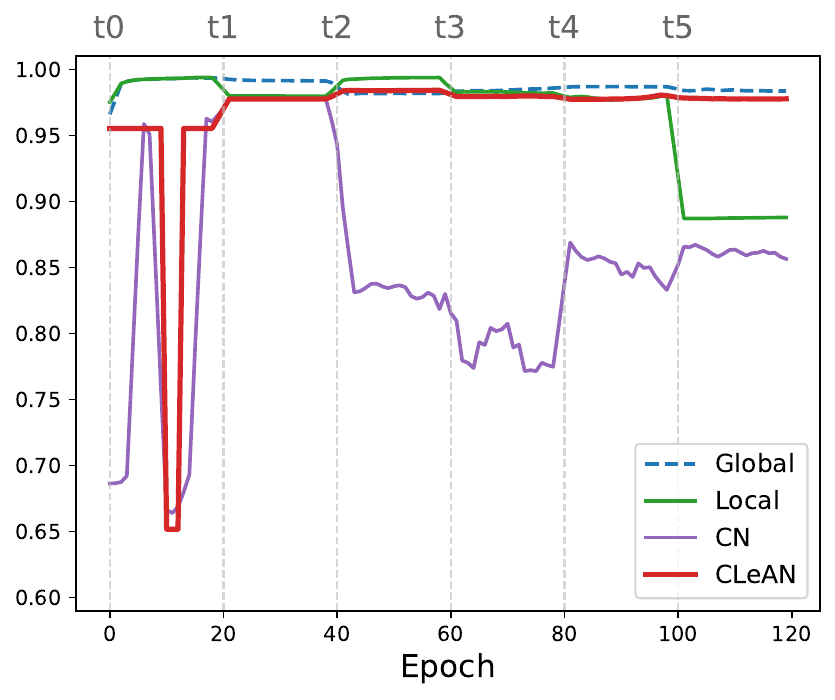} &
        \includegraphics[width=0.22\linewidth]{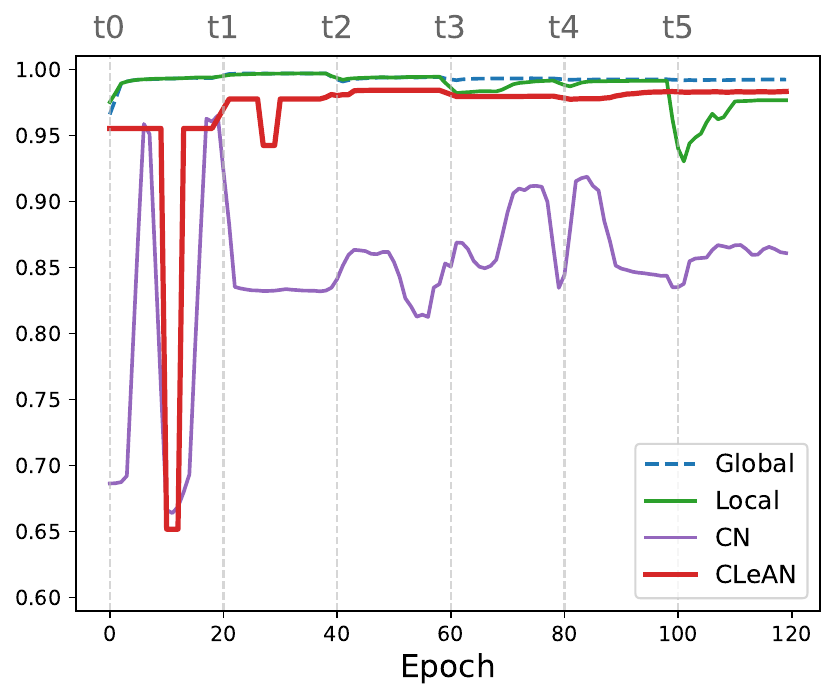} &
        \includegraphics[width=0.22\linewidth]{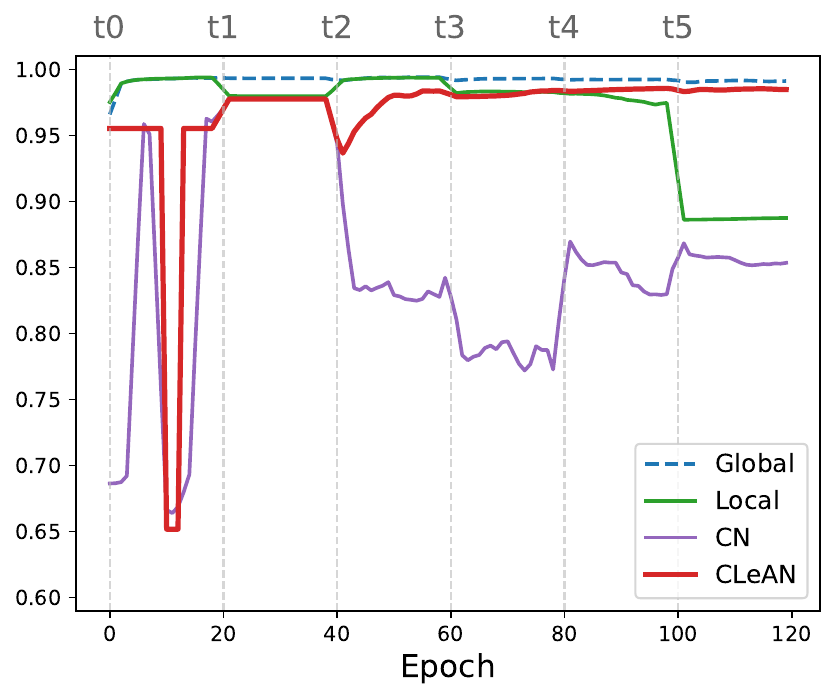} &
        \includegraphics[width=0.22\linewidth]{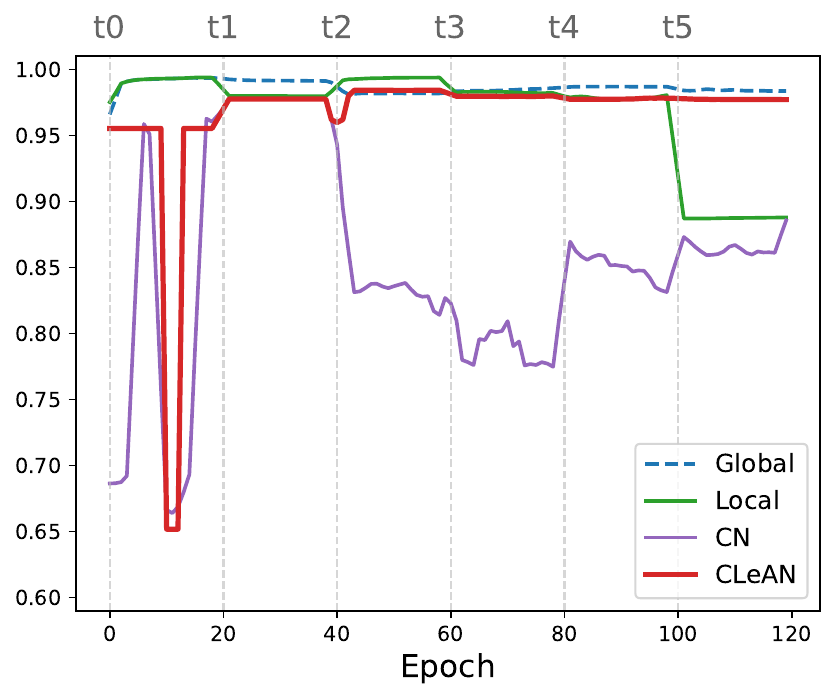} \\
        & (a) & (b) & (c) & (d) \\
        \rotatebox{90}{\hspace{18px}\textbf{Average AUROC ($\uparrow$)}} &
        \includegraphics[width=0.22\linewidth]{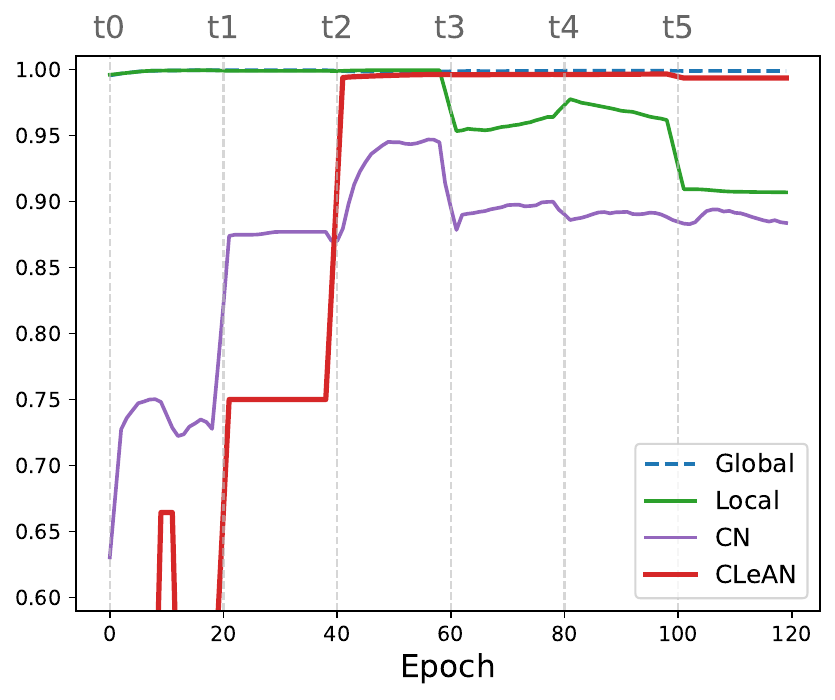} &
        \includegraphics[width=0.22\linewidth]{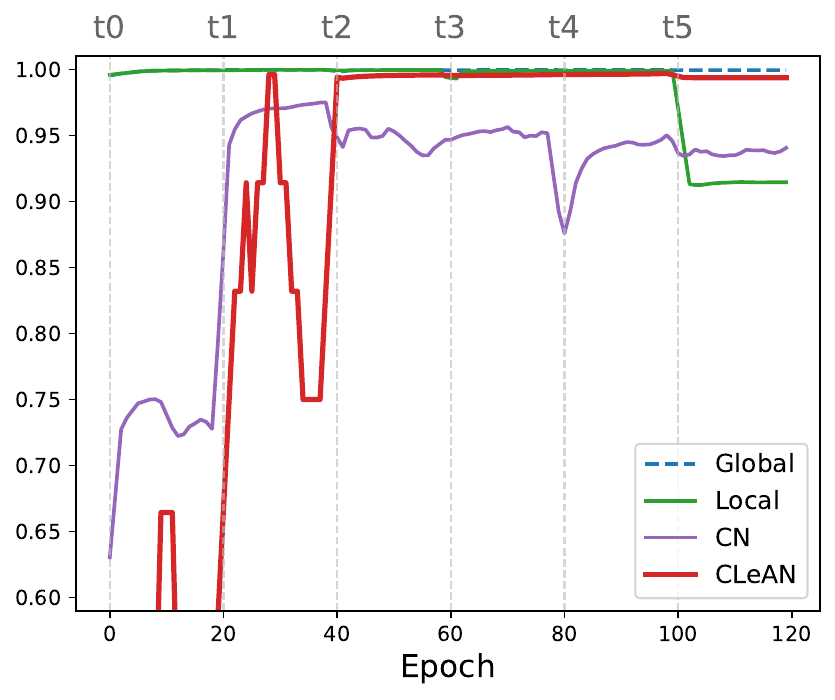} &
        \includegraphics[width=0.22\linewidth]{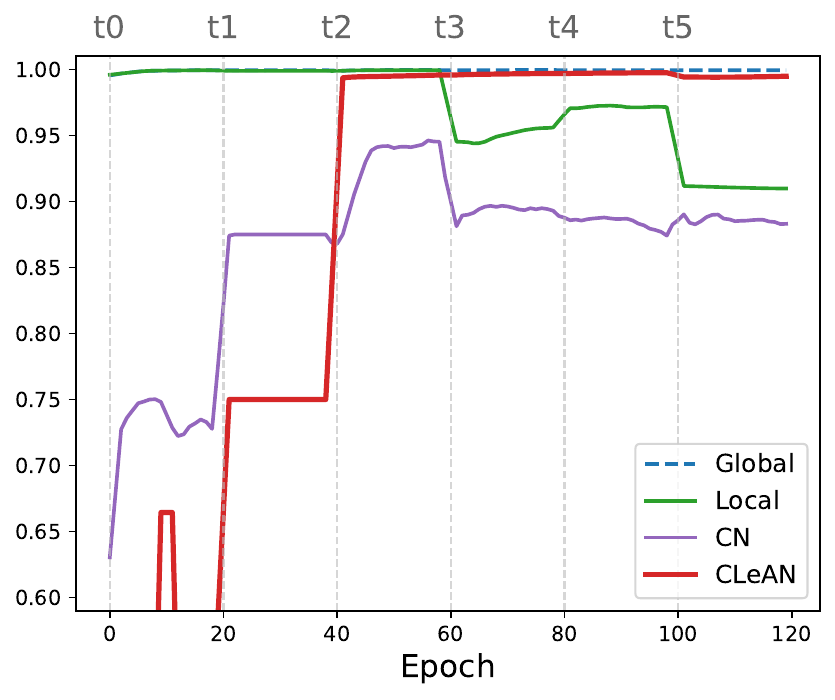} &
        \includegraphics[width=0.22\linewidth]{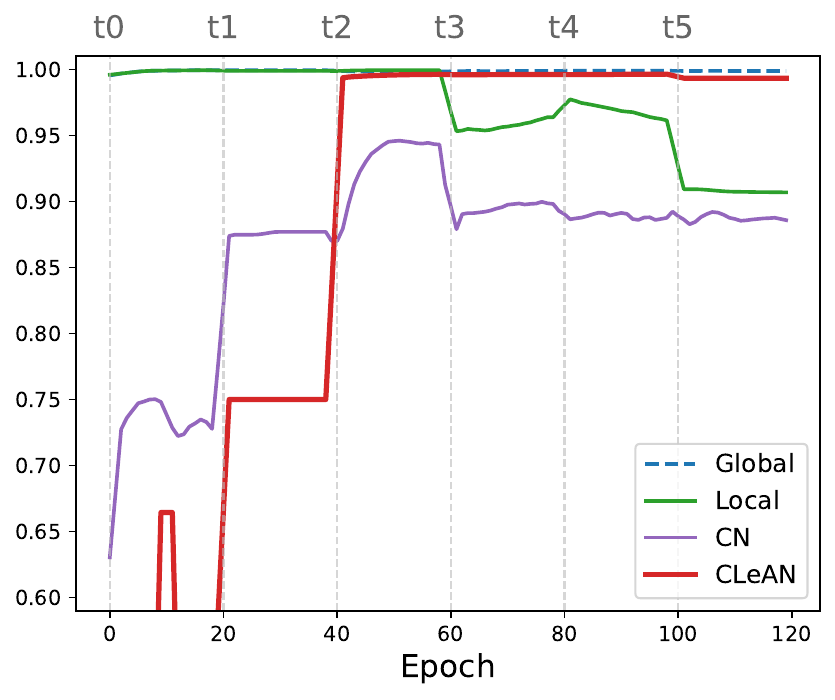} \\
        & (e) & (f) & (g) & (h) \\
        \rotatebox{90}{\hspace{18px}\textbf{Average Forgetting ($\uparrow$)}} &
        \includegraphics[width=0.22\linewidth]{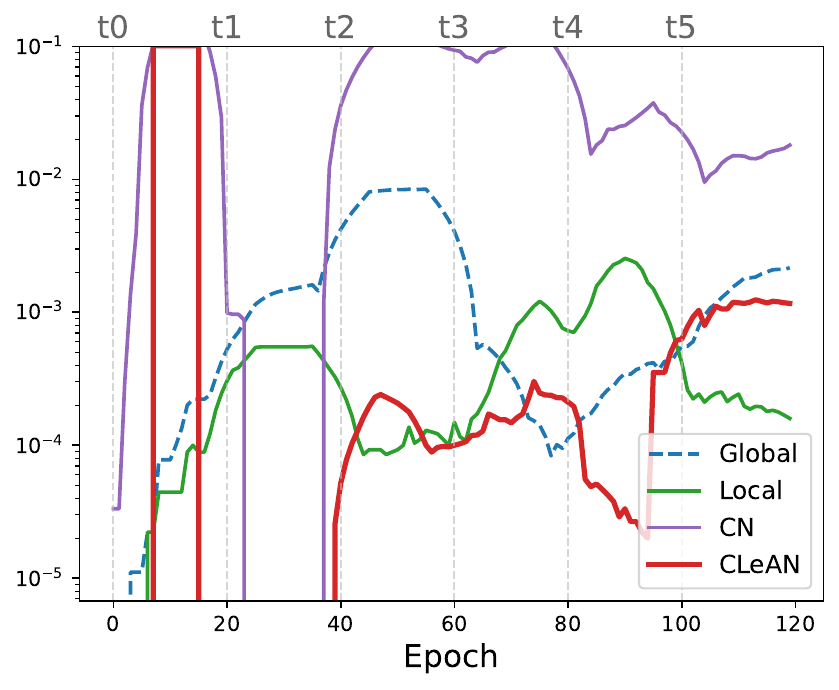} &
        \includegraphics[width=0.22\linewidth]{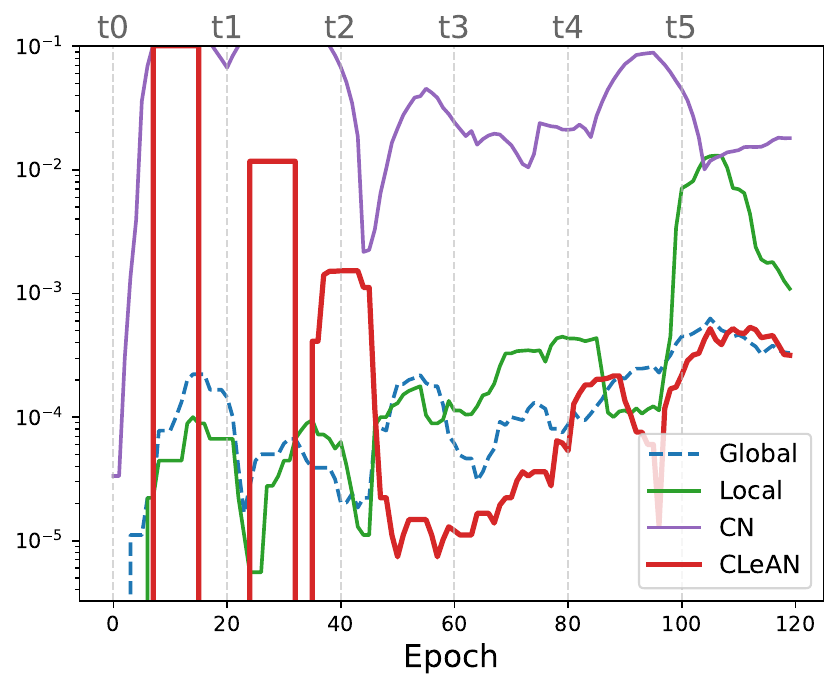} &
        \includegraphics[width=0.22\linewidth]{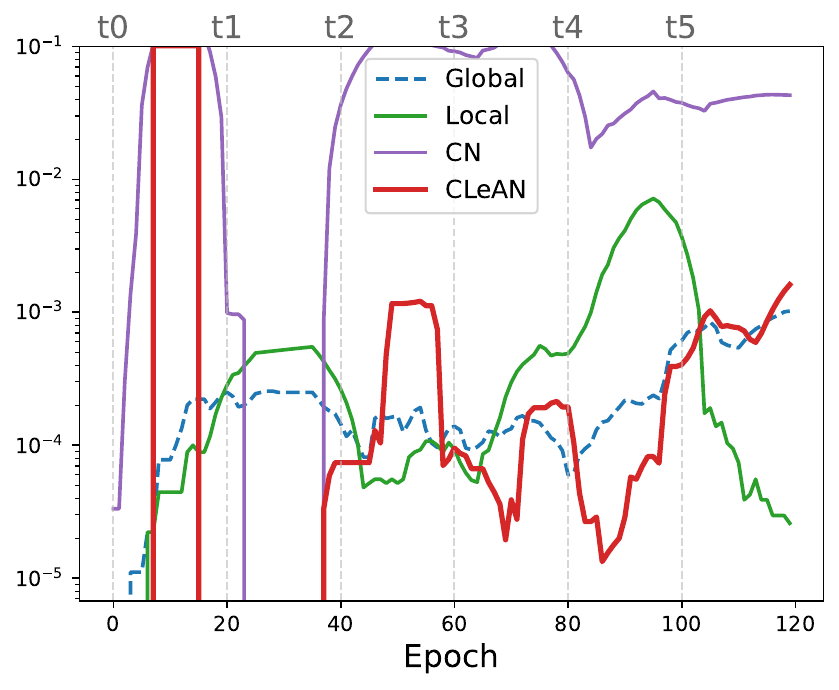} &
        \includegraphics[width=0.22\linewidth]{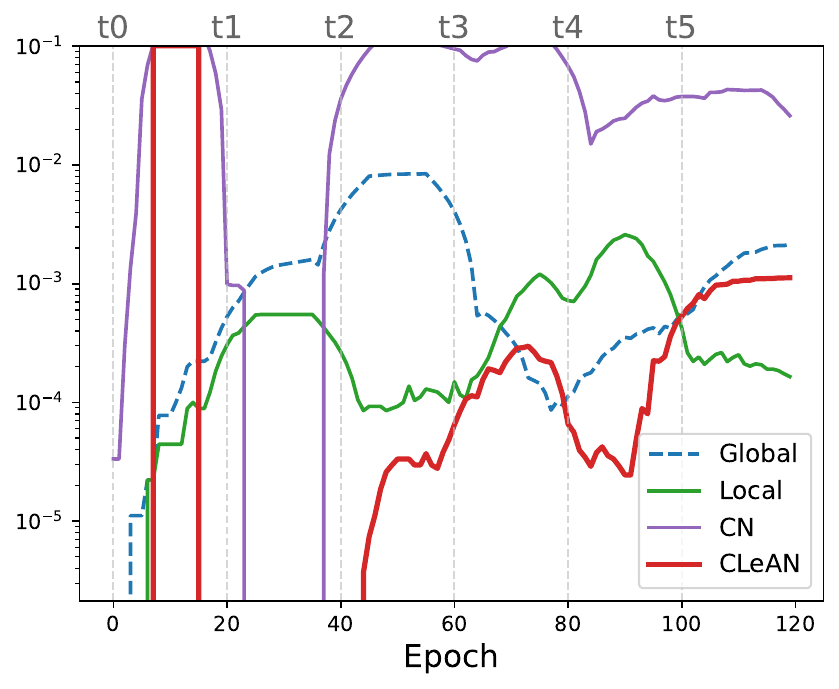} \\
        & (i) & (j) & (k) & (l) \\
    \end{tabular}
    \captionof{figure}{Performance comparison on the UNSW-NB15 dataset. 
    \textit{(a–d)} Average Accuracy, \textit{(e–h)} Average AUROC, and \textit{(i–l)} Average Forgetting are shown across the five daily experiences for different buffer strategies: 
    \textit{(a, e, i)} Finetuning, \textit{(b, f, j)} Reservoir, 
    \textit{(c, g, k)} A-GEM, and \textit{(d, h, l)} EwC. 
    Each plot compares the results obtained with different normalization techniques (Global, Local, CN, CLeAN).}
    \label{fig:results_UNSW-NB15}
\end{table*}

%%%%%%%%%%%%%%%%%%%%%%%%%%%%%%%%%%%%%%%%
\subsection{Results}
We discuss the results related to the UNSW-NB15 and CICIDS-2017 datasets. The outcomes are summarized in \Cref{fig:results_UNSW-NB15} and \Cref{fig:results_CICIDS-2017}, which report, for each dataset and each continual learning strategy (Finetuning, Reservoir Experience Replay, A-GEM, and EWC), the three metrics of interest introduced in~\Cref{ssec:metrics}: average accuracy, average forgetting, and average AUROC. Each figure is structured in rows, with one row per metric, and in columns, with one column per continual learning strategy. Within each plot, all normalization methods are displayed simultaneously, allowing direct visual comparison of their behavior over time. The shaded vertical separators highlight the transitions between experiences, while curves are plotted across the entire training span of each experience. This layout allows us to interpret how each combination of normalization method and continual learning strategy evolves throughout the sequential data stream, both in terms of learning capability on new data and retention of previously acquired knowledge.

Across all experiments, the impact of normalization emerges as a central factor influencing both accuracy and catastrophic forgetting. The difference between normalization schemes is evident from the earliest phases of training and persists throughout the data stream, confirming that an appropriate normalization strategy is essential when learning from tabular data under continual distribution shifts.

For comparative purposes, we report the \emph{global normalization} that represents a theoretical upper bound in terms of accuracy and AUROC for all realistic approaches because it exploits information from the entire dataset including future data. %At the opposite end, the absence of normalization leads to the poorest results. 
In \Cref{fig:results_UNSW-NB15}, we can see that the model rapidly saturates around relatively low accuracy values (approximately $80\%$) and exhibits severe forgetting. This is consistent with the significant variability in scale, where several features assume values in the order of units while others reach magnitudes of millions. 

The behavior of \emph{CN} is also instructive. Although CN was originally designed for computer vision, where feature distributions tend to evolve smoothly over time, its performance in the tabular setting is considerably weaker. Both \Cref{fig:results_UNSW-NB15} and \Cref{fig:results_CICIDS-2017} show that CN produces curves that resemble the Finetuning baseline, with limited accuracy and substantial forgetting. This suggests that the dynamics of tabular network data, characterized by abrupt shifts, heavy-tailed distributions, and outliers, are not well handled by CN’s momentum-based adaptation of mean and variance.
\begin{table*}[thb]
    \centering
    \begin{tabular}{ccccc}
        & \texttt{Finetuning} & \texttt{Reservoir} & \texttt{A-GEM} & \texttt{EwC} \\
        \rotatebox{90}{\hspace{18px}\textbf{Average Accuracy ($\uparrow$)}} &
        \includegraphics[width=0.22\linewidth]{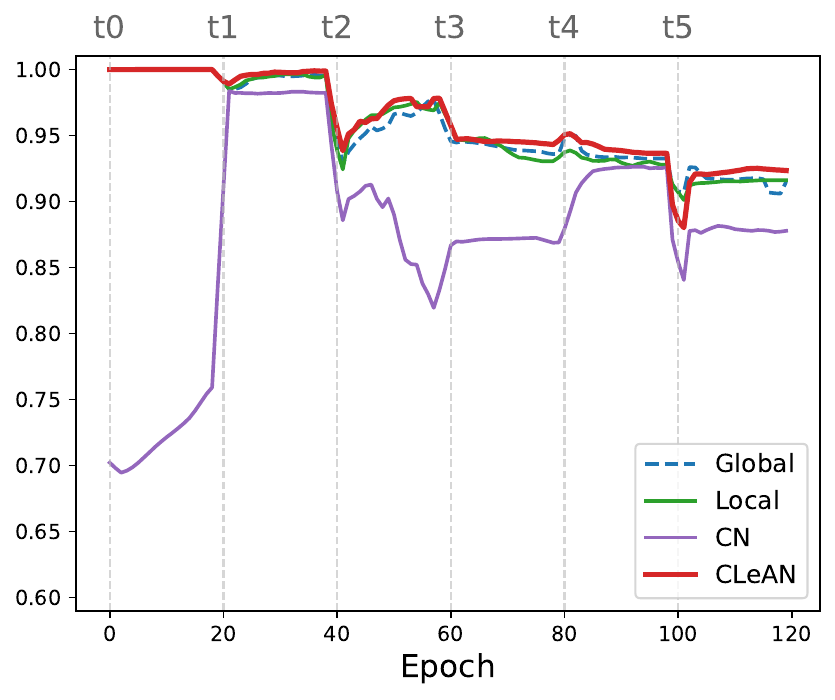} &
        \includegraphics[width=0.22\linewidth]{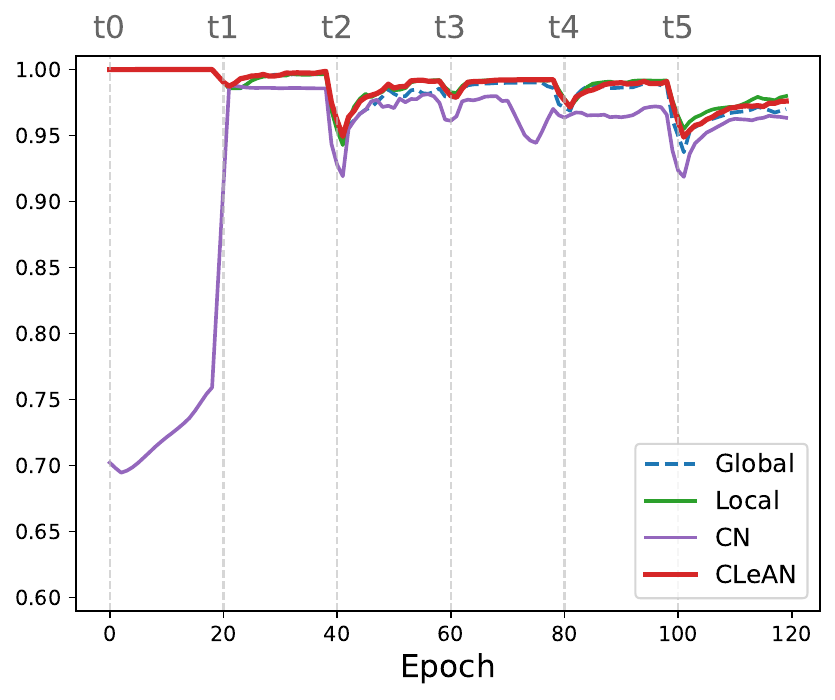} &
        \includegraphics[width=0.22\linewidth]{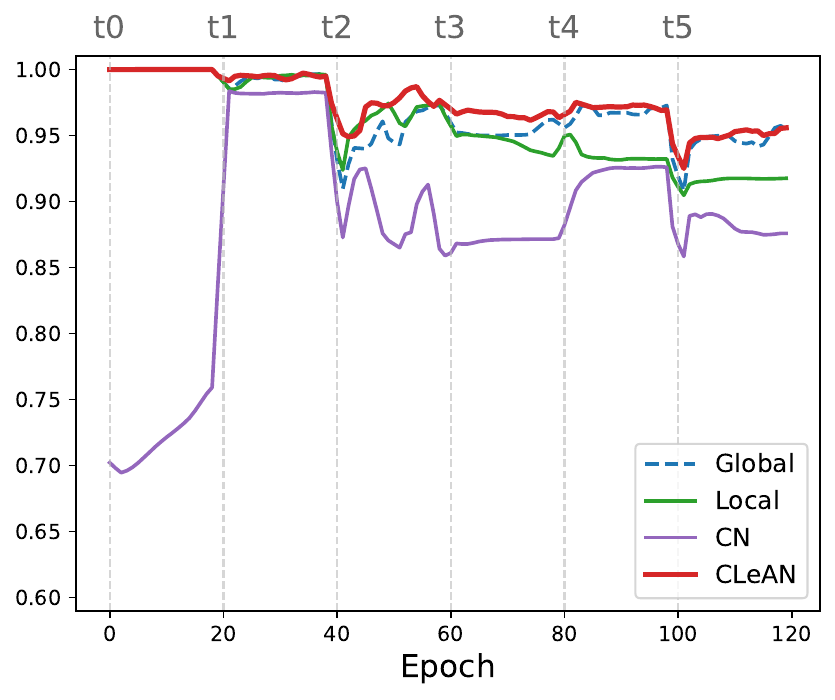} &
        \includegraphics[width=0.22\linewidth]{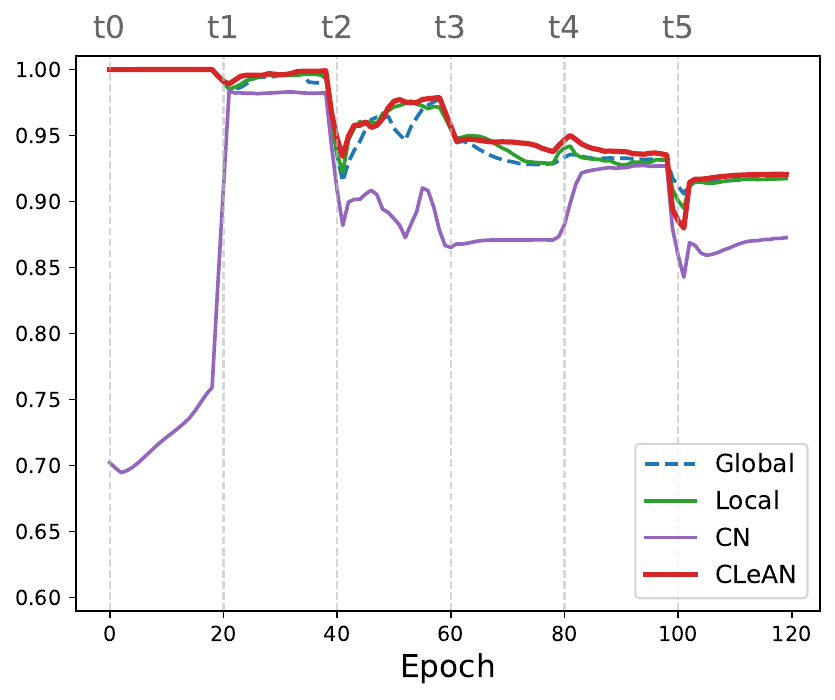} \\
        & (a) & (b) & (c) & (d) \\
        \rotatebox{90}{\hspace{18px}\textbf{Average AUROC ($\uparrow$)}} &
        \includegraphics[width=0.22\linewidth]{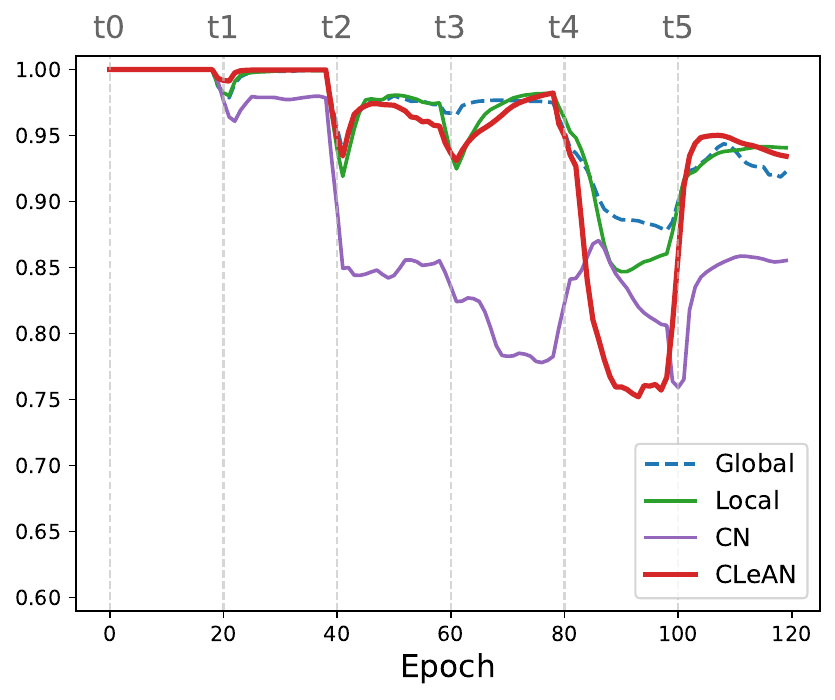} &
        \includegraphics[width=0.22\linewidth]{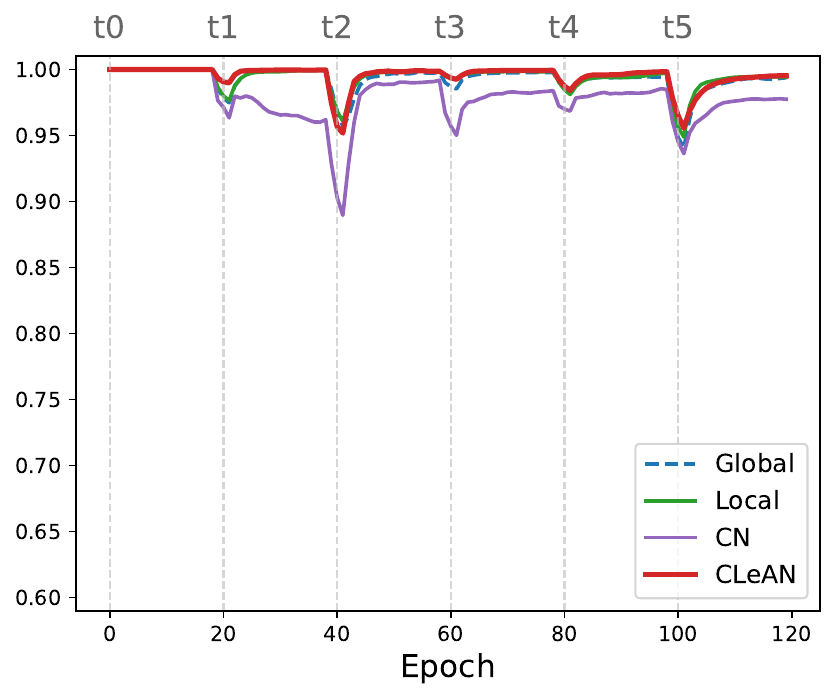} &
        \includegraphics[width=0.22\linewidth]{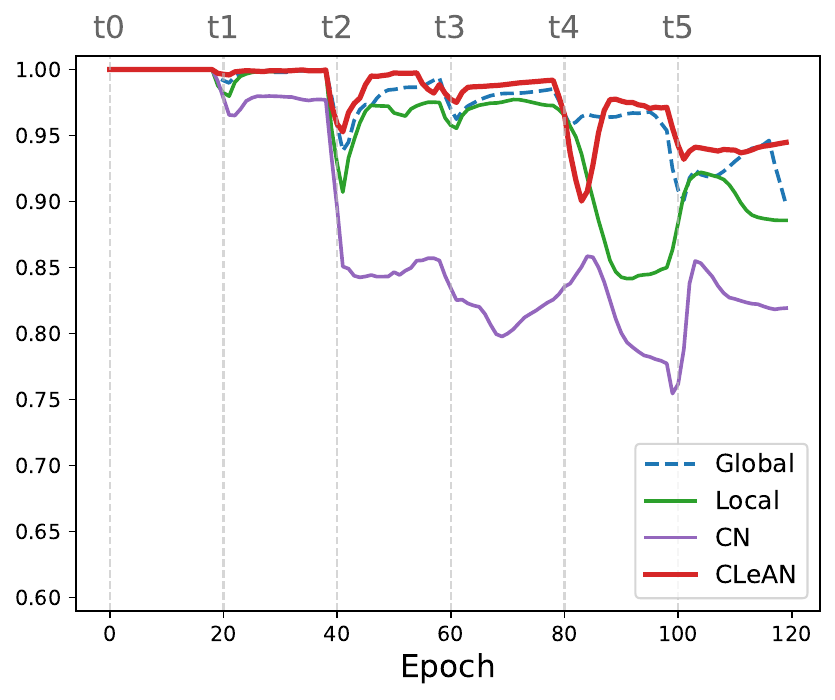} &
        \includegraphics[width=0.22\linewidth]{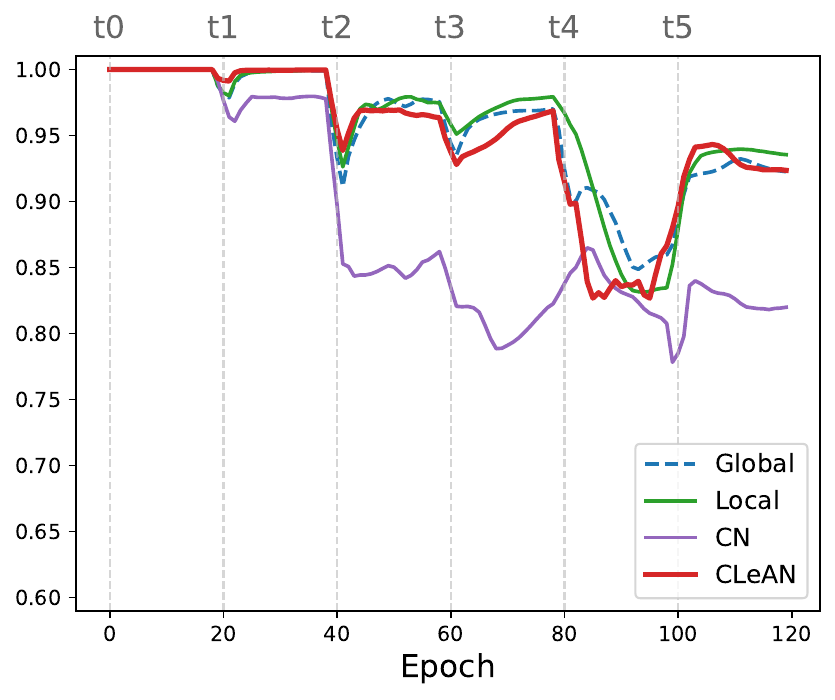} \\
        & (e) & (f) & (g) & (h) \\
        \rotatebox{90}{\hspace{18px}\textbf{Average Forgetting ($\uparrow$)}} &
        \includegraphics[width=0.22\linewidth]{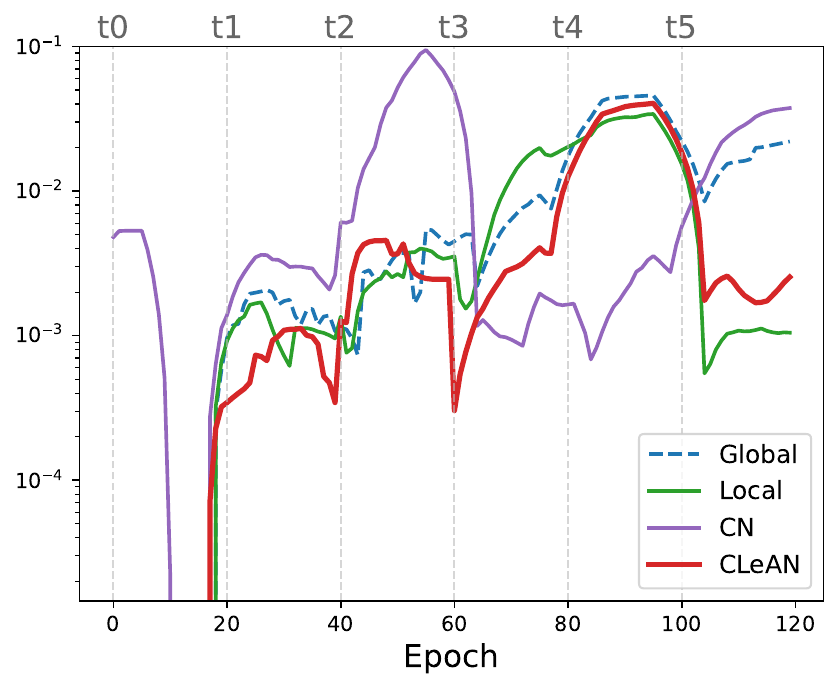} &
        \includegraphics[width=0.22\linewidth]{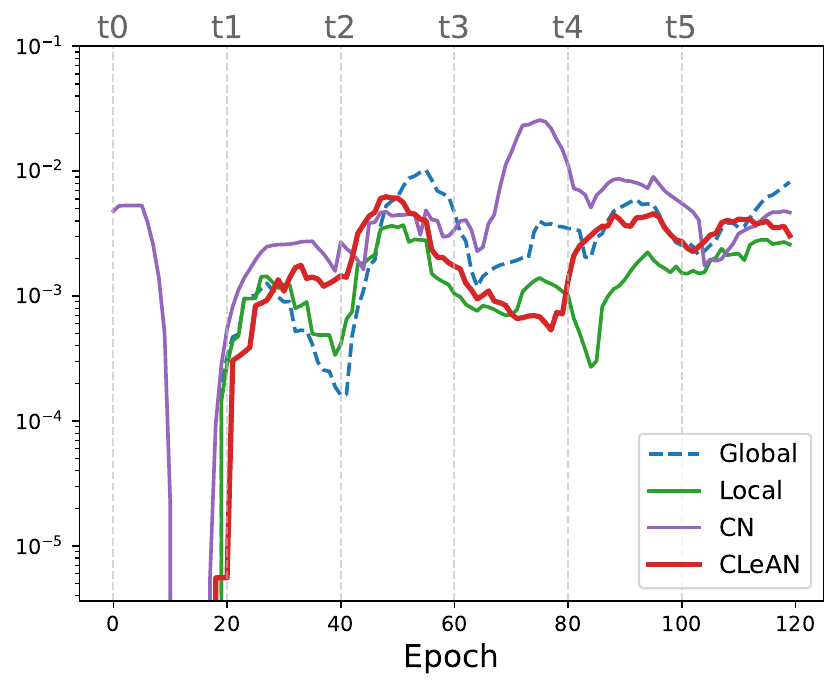} &
        \includegraphics[width=0.22\linewidth]{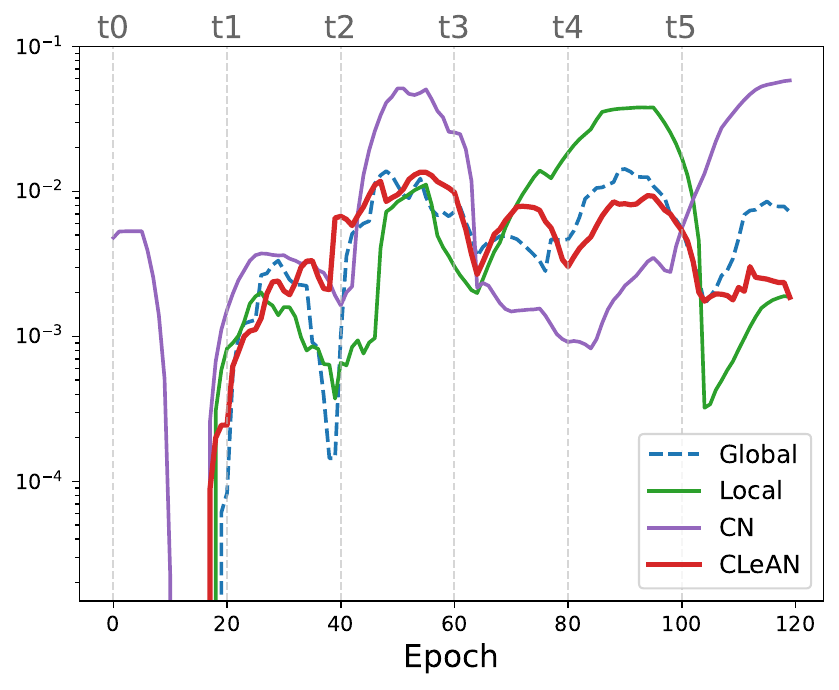} &
        \includegraphics[width=0.22\linewidth]{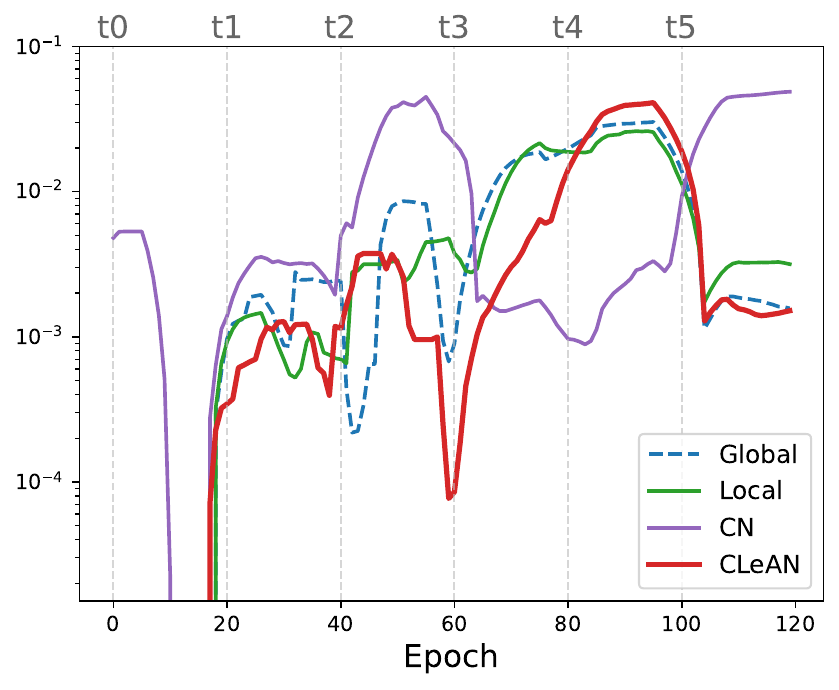} \\
        & (i) & (j) & (k) & (l) \\
    \end{tabular}
    \captionof{figure}{Performance comparison on the CICIDS-2017 dataset. \textit{(a-d)} Average Accuracy, \textit{(e-h)} Average AUROC, and \textit{(h-l)} Average Forgetting are shown across the five daily experiences for different buffer strategies: \textit{(a, e, i)} Finetuning, \textit{(b, f, j)} Reservoir, \textit{(c, g, k)} A-GEM, and \textit{(d, h, l)} EwC. Each plot compares the results obtained with different normalization techniques (Global, Local, CN, CLeAN).}
    \label{fig:results_CICIDS-2017}
\end{table*}

The results of \emph{local normalization} are surprisingly appreciable especially in the early stages of training. By rescaling each chunk independently, it can mitigate the scale imbalance problem and enables the network to learn effectively during the first experiences. In the UNSW-NB15 experiments (\Cref{fig:results_UNSW-NB15}), local normalization benefits  from Reservoir Experience Replay, where it maintains high accuracy across all experiences and exhibits reduced forgetting. Conversely, its behavior for EWC is  similar to finetuning. This  indicates that purely regularization-based approaches do not compensate for the variability introduced by local min–max scaling.

The proposed method, CLeAN, consistently achieves the best results that, after a brief stabilization phase, reach the theoretically optimal curve for accuracy and AUROC on both datasets. Its main advantage lies in its ability to preserve past knowledge: the average forgetting is lower than that of most other methods, particularly in the later experiences where distribution shifts are more severe. This can be observed in \Cref{fig:results_UNSW-NB15}(i–l) and \Cref{fig:results_CICIDS-2017}(i–l), where the CLeAN curve remains consistently lower than all other strategies except local normalization, which, in most approaches, yields lower average forgetting than CLeAN. This apparent discrepancy is explained by the definition of forgetting itself. A method that achieves low accuracy may also exhibit artificially low forgetting, since degradation is measured relative to its best past performance. CLeAN exhibits a poor behavior in the initial two experiences. This early dip is expected and motivated because during the initial phases the scaling network $s_\Phi$ is still untrained and the EMA module has not yet stabilized the estimates of the global min–max statistics. As these components converge, the normalization becomes more reliable, and the performance rapidly increases, aligning with the trends observed in the later experiences. This behavior could be partially controlled through the smoothing parameter $\eta$ of the EMA module: smaller values of $\eta$ would allow the EMA to adapt more quickly and to rely earlier on the predictions of the normalization model $s_\Phi$. We have also to consider that in intrusion detection scenarios, early-stage performance is less critical than stability and accuracy in later stages, where distribution shifts are more pronounced. From this perspective, the initial performance dip is not a limitation but rather a desirable feature, as it allows the normalization process to stabilize before fully trusting learned statistics. A systematic analysis of the effect of $\eta$ is left for future work, as it falls outside the scope of this paper.

The experiments on CICIDS-2017 (\Cref{fig:results_CICIDS-2017}) lead to the same conclusions. Here, the first experience contains only benign samples, which explains why all methods reach near-maximal accuracy at $t_0$. Nevertheless, as soon as malicious traffic appears in $t_1$, the differences between normalization schemes become prominent. CLeAN combined with Reservoir Experience Replay emerges as the most effective configuration, reaching high accuracy and AUROC while keeping forgetting remarkably low. In contrast, EWC consistently performs as one of the least effective continual learning strategies, often matching the behavior of finetuning, further emphasizing the challenges posed by tabular data.

Overall, the results highlight the crucial role of normalization in continual learning for tabular data. CLeAN not only stabilizes training and facilitates rapid adaptation to evolving data distributions, but also acts as a powerful mechanism for reducing forgetting. Its performance across different datasets and continual learning methods demonstrates that adaptive normalization should be regarded as a core component of any continual learning pipeline for tabular network data.

%%%%%%%%%%%%%%%%%%%%%%%%%%%%%%%%%%%%%%%%%%%
% CONCLUSION
%%%%%%%%%%%%%%%%%%%%%%%%%%%%%%%%%%%%%%%%%%%
\section{Conclusions}\label{conclusions}
This work addressed a fundamental yet largely overlooked challenge in continual learning for tabular data: the role of data normalization under sequential, non-stationary data streams. While normalization is typically treated as a static preprocessing step, our analysis shows that this assumption breaks down in realistic scenarios, such as cybersecurity, where feature distributions may shift abruptly due to protocol changes, evolving traffic patterns, or the emergence of new attack classes. We demonstrated empirically that traditional strategies such as global min-max scaling or local per-chunk normalization not only impact performance but can also substantially alter the extent of catastrophic forgetting, underscoring normalization as a core determinant of stability in continual learning.

To overcome the limitations of existing approaches, we introduced CLeAN, a learnable and adaptive normalization mechanism that estimates the global feature range using an Exponential Moving Average and refines it through a lightweight scaling network. Unlike global normalization, CLeAN operates under realistic streaming constraints, and unlike local normalization or continual normalization, CLeAN achieves robustness to abrupt distribution shifts. Across two benchmarks based on IDS datasets and multiple continual learning strategies (Finetuning, Reservoir Experience Replay, A-GEM, and EWC), CLeAN consistently achieves accuracy and AUROC values close to those of global normalization, at least after stabilization of the EMA module and of the normalization model $s_\Phi$, while substantially reducing catastrophic forgetting. In several configurations, CLeAN even enables improved performance on past experiences, suggesting that adaptive normalization can act as a stabilizing force for long-term memory.

Beyond its empirical benefits, this work highlights a broader conceptual insight: normalization should not be regarded as a preprocessing step, but as an integral component of the continual learning pipeline. Our findings suggest that adaptive normalization mechanisms may complement or enhance existing replay- and regularization-based methods, offering a new avenue for improving resilience to distribution drift in tabular data.

Future work will investigate how CLeAN interacts with more advanced continual learning techniques, including hybrid replay-regularization strategies and uncertainty-aware models. Extending adaptive normalization to multimodal scenarios or deploying CLeAN within large-scale, real-time systems represents a promising direction for practical adoption. Overall, this study demonstrates that rethinking normalization provides a principled and effective path toward more robust and adaptive systems in dynamic environments.

\bibliographystyle{IEEEtran}
\bibliography{bibliography}
\end{document}